\documentclass[letterpaper, 10 pt, conference]{ieeeconf}  % Comment this line out if you need a4paper

\IEEEoverridecommandlockouts                              % This command is only needed if 
\usepackage{graphics} % for pdf, bitmapped graphics files
\usepackage{amsmath} % assumes amsmath package installed
\usepackage{amssymb}  % assumes amsmath package installed

\usepackage{makecell}

\usepackage{xcolor,colortbl}
\usepackage{hyperref}
\usepackage{float}
\usepackage{graphicx}
\usepackage{pifont}
\usepackage{bbold}

\usepackage{makecell}
\usepackage{algorithm}
\usepackage{algpseudocode}

\algnewcommand\algorithmicforeach{\textbf{for each}}
\algdef{S}[FOR]{ForEach}[1]{\algorithmicforeach\ #1\ \algorithmicdo}

\definecolor{Gray}{gray}{0.95}

\usepackage{tabularx}
    \newcolumntype{L}{>{\centering\arraybackslash}X}

\newcommand{\OA}[1]{} %\par\textcolor{purple}{[OA: #1]}\par}

\newcommand{\METHOD}{AssembleRL} %DRL4ASSEMBLE} %Erl: Bunu daha sonra degistirelim
\newcommand{\COMP}{\mu_{cmp}} % InCompleteness
\newcommand{\CORR}{\mu_{cor}} % InCorrectness

\title{\LARGE \bf
AssembleRL: Learning to Assemble Furniture \\ from Their Point Clouds
}

\author{Ozgur Aslan$^{1}$, Burak Bolat$^{1}$, Batuhan Bal$^{1}$, Tugba Tumer$^{1}$, Erol Sahin$^{1}$, Sinan Kalkan$^{1}$% <-this % stops a space
\thanks{*Partially supported by TUBITAK with projects 120E269 and 117E002. }% <-this % stops a space
\thanks{$^{1}$All authors are with KOVAN Research Lab; METU-ROMER, Center for Robotics and Artificial Intelligence; and Dept. of Computer Engineering, Middle East Technical University, Ankara, Turkey.
        {E-mail: aslan.ozgur@metu.edu.tr}}
}

\begin{document}

%% TODO
% - completeness vs. incompleteness?
%
% - Baslik ve isim finalize 
% - Sekiller [Erol Hoca kagit uzerinde cizecek]
% - Yontem [Erol]
%   + Reward pseudocode with algorithm + algorithmic
% - Deneyler
%  + Reward & measure grafikleri [Ozgur]
%  + Ablation on distance thresholds 
%  + Tablo 2'ye complexity eklenecek
%  + Ablation without anchor [PAPER SONRASI]

\maketitle
\thispagestyle{empty}
\pagestyle{empty}

%%%%%%%%%%%%%%%%%%%%%%%%%%%%%%%%%%%%%%%%%%%%%%%%%%%%%%%%%%%%%%%%%%%%%%%%%%%%%%%%
\begin{abstract}
The rise of simulation environments has enabled learning-based approaches for assembly planning, which is otherwise a labor-intensive and daunting task. Assembling furniture is especially interesting since furniture are intricate and pose challenges for learning-based approaches. Surprisingly, humans can solve furniture assembly mostly given a 2D snapshot of the assembled product. Although recent years have witnessed promising learning-based approaches for furniture assembly, they assume the availability of correct connection labels for each assembly step, which are expensive to obtain in practice. In this paper, we alleviate this assumption and aim to solve furniture assembly with as little human expertise and supervision as possible. To be specific, we assume the availability of the assembled point cloud, and comparing the point cloud of the current assembly and the point cloud of the target product, obtain a novel reward signal based on two measures: Incorrectness and incompleteness. We show that our novel reward signal can train a deep network to successfully assemble different types of furniture. Code and networks available here: https://github.com/METU-KALFA/AssembleRL
\end{abstract}

\section{Introduction}
\label{sect:introduction}

\begin{figure}[H]
    \centering
    \includegraphics[width=0.9\linewidth]{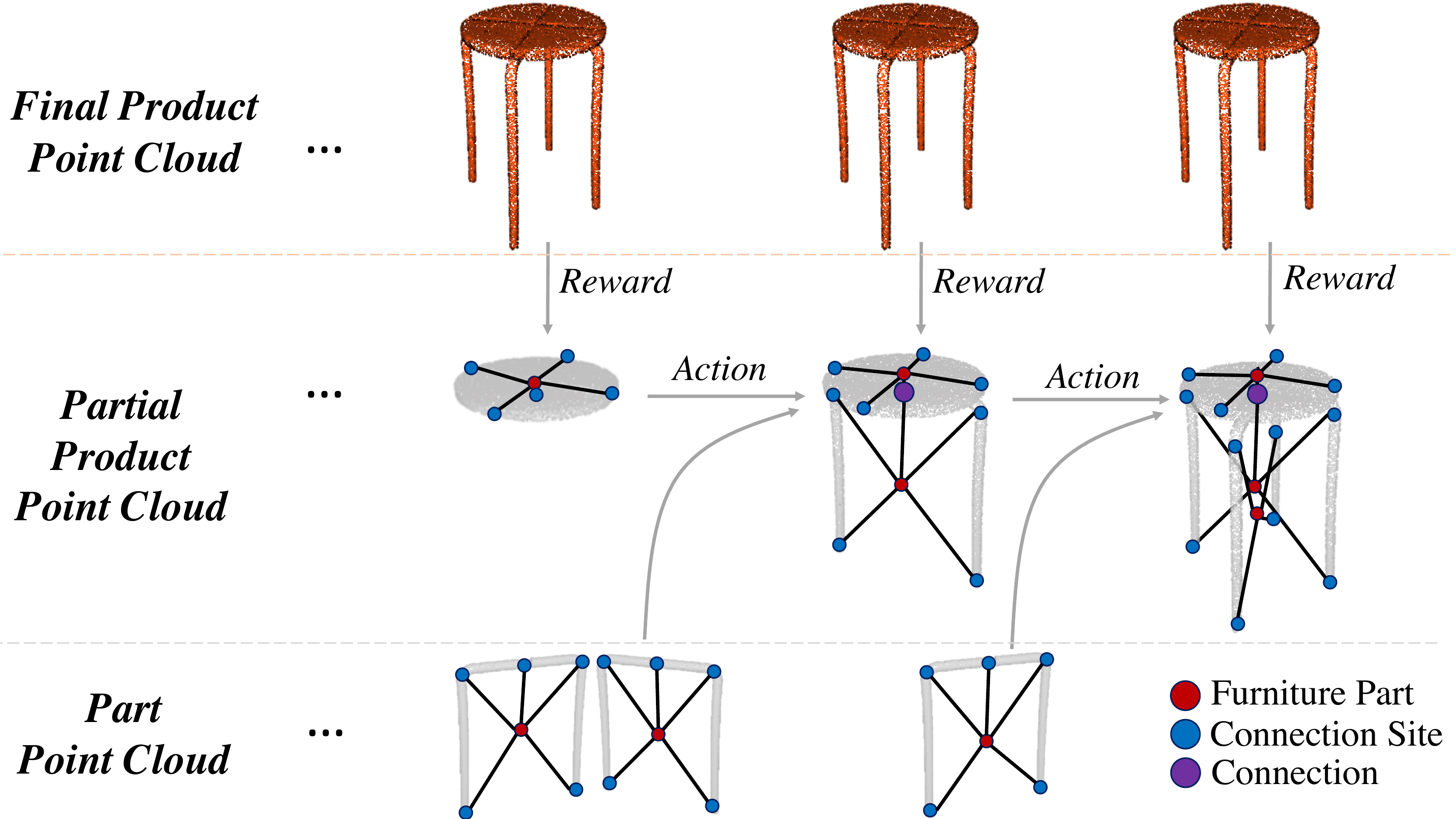}
    \caption{We propose learning to assemble a product given only its assembled form, unlike prior work that requires precise supervision for each assembly step. Our novel reward formulation uses two measures; one for measuring the incorrectness of the assembly so far, and another for the incompleteness of the assembly with respect to the target (see Fig. \ref{fig:measurement}).
    \label{fig:teaser}}
\end{figure}

Assembling a product from its parts is a challenging task that fascinates kids as well as adults who prefer to build their toys/furniture from parts provided in LEGO/IKEA boxes. Although a sequence of line drawings is provided as the assembly plan of these products, the final view of the assembled product often serves as the ultimate guide, enabling humans to fill in the gaps in the assembly plans by comparing the partially assembled product with the final version. Such an assembly capability is desirable for robots to be deployed in low-volume assembly tasks, where the overhead of specifying a detailed assembly plan takes away the benefits of automation. In this sense, we are motivated by the vision of  a ``Assembly Robot'' that you would rent to build your boxed furniture on your behalf.

%To this end, for each product, a human expert using necessary tools creates a rigorous assembly plan (e.g.  \cite{10.1145/882262.882352} and \cite{assembly_plan}), a motion plan (e.g. \cite{intmotionplanning}) and an optional sensing system (e.g. \cite{SANTOCHI1998503}) to detect failures. However, this ``design'' phase is very labour intensive and prone to errors owing to the human expert. 

In this paper, we propose a Deep Reinforcement Learning (DRL) based method, \METHOD, to learn assembly plans using the final view of the assembled furniture as a guide, along with the specifications and view of its parts (Fig. \ref{fig:teaser}). Our work, along with prior studies~\cite{2_RoboAssembly, 4_Learn2Assembly} on learning-based furniture assembly, are motivated by the availability of the IKEA furniture assembly simulation environment~\cite{lee2021ikea} which also includes a library of furniture models. 

%on furniture assembly due to the availability of IKEA furniture assembly simulation environment \cite{lee2021ikea}, along with the variety of furniture models. 

%% Erl: Bu paragraf guzel ama daha sonraya kaydiralim.
%The rise of learning-based approaches, especially deep reinforcement learning, have led to unprecedented applications for programming and controlling robots. For example, robots can be programmed to walk, fly, perform manipulation tasks or navigate in challenging environments (see e.g. \cite{doi:10.1177/17298814211007305} for a review). In developing learning-based approaches for such robotic applications, as it is difficult to collect sufficient amounts of data in the real world, realistic and physics-based simulation environments are generally used.

%In~\cite{2_RoboAssembly}   Yu et al. used Reinforcement Learning to estimate pairs to be assembled with a connection site at each step. Huang et al. \cite{1_3Dgen}, on the other hand, proposed a 6D pose estimation for each part using Graph Neural Network (GNN). These studies, however, assume that part poses at the target model are known and a segmented target model is given. %%Erl Bu ozetler yapilmis olan isleri tam anlatmiyor

In this paper, in contrast to prior work \cite{2_RoboAssembly}, we propose to use only the fully assembled point cloud of the furniture and the mesh models of its parts, to learn the assembly plan. Specifically, we introduce a novel reward function that evaluates the match between the point cloud of the partially assembled furniture against its fully assembled view using two measures that evaluate the {\em incorrectness} and {\em incompleteness}. We train a graph-convolutional neural network with our novel reward signal, combining the incorrectness and incompleteness measures, to learn the assembly plan as a policy that predicts which part pairs need to be connected via which of their connections. The method is successfully tested on 11 IKEA furniture models.

%using only the target 3D model for learning an assembly plan for furniture. To be specific, we introduce a novel reward function that evaluates the match between the point cloud of the current assembly and the target product in terms of two measures: Correctness and completeness of the current assembly with respect to the target furniture. We show that we can train a graph-convolutional neural network using our novel reward signal and apply our solution for learning 10 different furniture.

Our main contributions are: (1) We only use the target point cloud to learn assembly plans. (2) We introduce a novel reward function that quantifies incorrectness and incompleteness of the current assembly with respect to the target model. (3) We apply our solution to learning assembly plans for different furniture.

\section{Related Work and Background}
\label{sect:related_work}
    
%\subsection{Assembly with Robots} 

Assembling with robots requires solving three main tasks \cite{jimenez2013survey, support_tools}: Modelling, planning and execution. Modelling pertains to obtaining a representation of the assembly process and includes representing parts, tools, actions etc. A common approach for assembly modelling is using graphs (e.g. \cite{54734, 4_Learn2Assembly, 7_RGLNETAR, graphrep}), as they are naturally suitable for representing entities and the relations among them. 

Assembly planning is finding a sequence of assembly actions that, once executed, lead to the assembled product is an NP-complete problem \cite{kavraki1993complexity}. Although various backward or forward planners can be used for finding assembly plans, they generally require constraints about the task (provided by a human expert) and the resulting plan can be sub-optimal \cite{huang1991framework,lee1992backward,ghandi2015review}. Therefore, in practice, human experts are needed either for creating the whole assembly plan or for collaborative assembly planning/execution \cite{huang1991framework,rizwan2020human}. Automatic discovery of such plans, with little supervision, would benefit the development of ``Assembly Robots".

\subsection{Learning to Assemble Furniture}

The rise of learning-based approaches, especially DRL, has led to unprecedented success in programming and controlling robots (see e.g. \cite{doi:10.1177/17298814211007305} for a review), which have motivated such approaches for solving assembly tasks as well. As learning an assembly with a real robot can be costly, generally simulation environments are used by learning-based assembly approaches \cite{2_RoboAssembly,lee2021ikea,robosuite2020}. %\cite{robosuite2020, lee2021ikea} focuses on sim-to-real transfer by providing accurate physics simulation with high quality rendering. Moreover, \cite{lee2021ikea, 2_RoboAssembly} focuses on furniture assembly. In our work, we propose a custom Gym environment that can be integrated with the above simulation environments.

%With the increasing popularity of deep reinforcement learning methods, there are works such as \cite{DBLP:journals/corr/abs-1803-07635} and \cite{DBLP:journals/corr/abs-1903-01066} that focus on learning robot control for an assembly task execution while learning a sequence implicitly. Different from these works, we focus on learning a policy that will generate the sequence of the assembly.

It has been shown that rotation and translation between object parts can be learned to assemble objects. For this purpose, deep networks such as Convolutional Neural Networks (CNN) \cite{3_SingleImage} or Graph Convolutional Networks (GCN) \cite{1_3Dgen} can be used. Such networks are trained using supervised learning with Chamfer Distance between the target and the assembled parts, L2 loss of the part translations and Chamfer Distance between whole assembled product with the target assembly as the supervision signal \cite{1_3Dgen} or RL using correct connection labels as the supervision signal \cite{2_RoboAssembly}.

The introduction of furniture assembly simulation environments \cite{2_RoboAssembly, lee2021ikea, robosuite2020} has enabled the use of learning-based approaches. %\cite{2_RoboAssembly}. %For example, Funk et al. \cite{4_Learn2Assembly} used GCNs and reinforcement learning to assemble boxes to match target 2D shape. 
For example, Huang et al. \cite{1_3Dgen} used supervised learning to train a graph neural network to estimate 6D pose for each part to assemble chairs, lamps and tables. 
Yu et al. \cite{2_RoboAssembly} employed DRL for chair assembly, though they assumed the availability of strong supervision (correct vs. incorrect labels for connections) for each assembly action.

\subsection{Comparative Summary}

Although there are promising learning-based approaches for furniture assembly, as listed in Table \ref{tab:comparison}, we see that they assume the availability of strong supervision (correct vs. incorrect connection labels) for each assembly step. In contrast, in this work, we only assume the availability of point cloud of the fully assembled furniture to provide weak supervision (reward) signal to train a DRL network. 

\begin{table}[hbt!]
    \centering
 \caption{Learning-based approaches for furniture assembly. \label{tab:comparison}}
 \scriptsize
    \begin{tabularx}{\linewidth}{|c|c|L|L|}
        \hline
        \textbf{Study} &  \textbf{Method} & \textbf{Weak-supervision} & \textbf{Output}\\
        \hline \hline
        \makecell{Huang et al.\\ \cite{1_3Dgen}} &  \makecell{GNN, \\ Supervised} & \ding{55} & \makecell{6D pose for \\each part}\\
        \hline
        \makecell{Yu et al.\\ \cite{2_RoboAssembly}} & RL & \ding{55}& \makecell{Pair,\\ Connection Site,\\ Rotation}\\ 
        \hline
        %Learn2Assembly & \ding{55} & GNN, RL & \ding{51} & Pair, placement side\\ \hline
        \textbf{\METHOD} &  GNN, RL & \ding{51}  & \makecell{Pair,\\ Connection Site}\\
        \hline
        
    \end{tabularx}
    
%\begin{table}[H]
%    \centering
% \caption{A summary of learning-based approaches for furniture assembly. \label{tab:comparison}}
%    \begin{tabularx}{\linewidth}{|c|L|c|L|L|}
%        \hline
%        Study & Furniture Assembly & Method & Weak-supervision & Output\\
%        \hline
%        Gen 3D & \ding{51} & \makecell{GNN, \\ Supervised} & \ding{55} & 6D pose for each part\\
%        \hline
%        RoboAssembly & \ding{51} & RL & \ding{55}& Pair, connection site, rotation\\ 
%        \hline
%        %Learn2Assembly & \ding{55} & GNN, RL & \ding{51} & Pair, placement side\\ \hline
%        Ours & \ding{51} & GNN, RL & \ding{51}  & Pair, connection site\\
%        \hline
%        
%    \end{tabularx}
   
\end{table}
\section{Methodology}
\label{sect:method}
%\textit{... vanilla PointNet that extracts a global permutation-invariant feature summarizing the input part point cloud ... We used shared PointNet...}
\subsection{Problem Definition}
\label{subsect:prob_def}
We define the problem as the discovery of an assembly plan using the  fully assembled point cloud view of the furniture, along with the connection specifications and the mesh models of its parts, which are sampled to obtain point cloud representations. Specifically, let us use $P^T$ to denote the point cloud of the fully assembled furniture, and $P^0$ \& $P^t$ to denote respectively the point cloud of the `seed' part at the beginning of the assembly, and the partially assembled furniture at step $t$ of the assembly. 

We assume that the assembly process starts with a single `seed' part, which is grown through the attachment of other parts towards the final product. Furniture that require the assembly of separate parts, such as assembly of drawers in a separate plan, which are then attached to the body of  a chest, are not addressed.
\begin{figure}[hbt!]
    \centering
    \includegraphics[width=\linewidth]{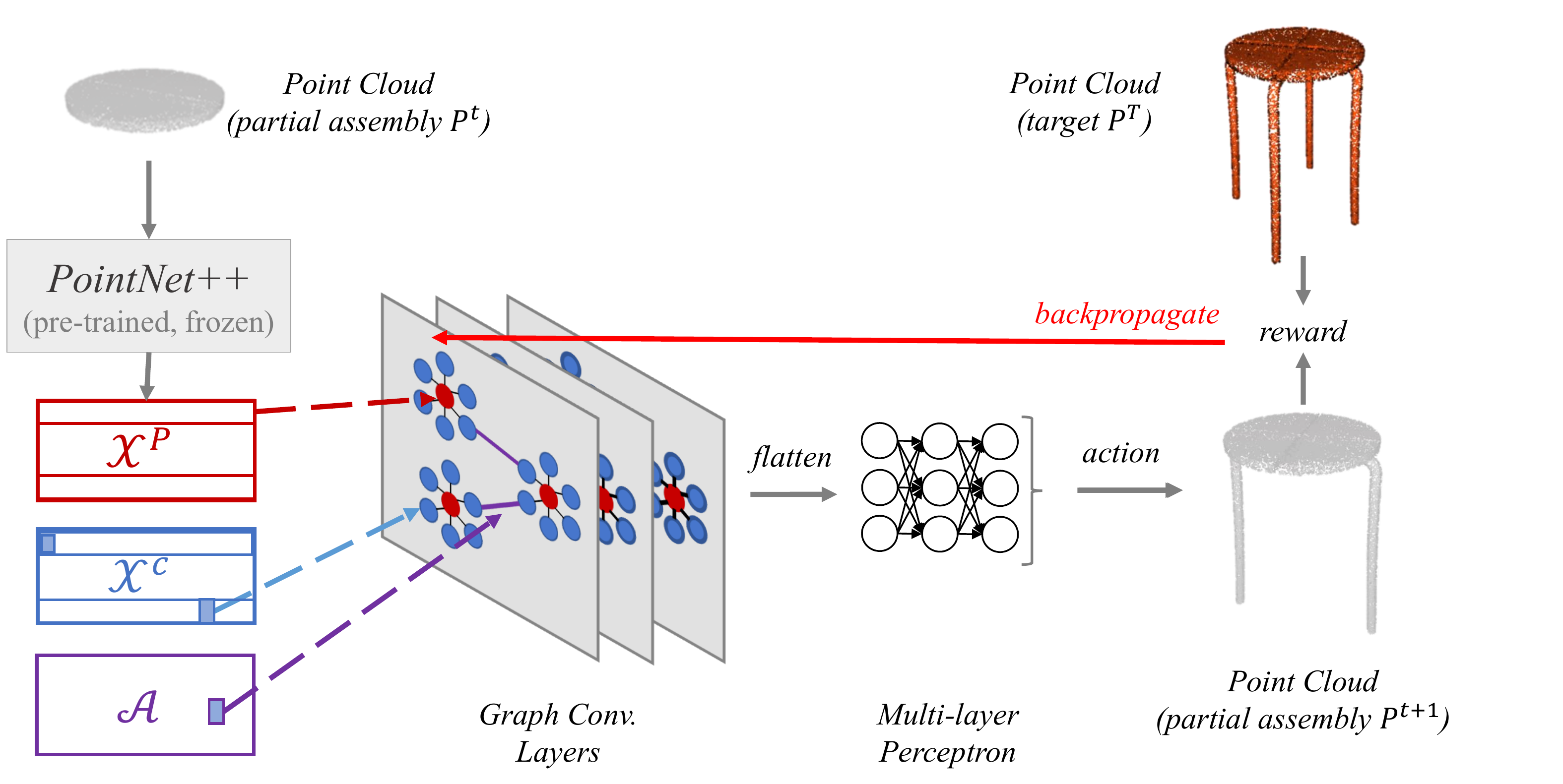}
    \caption{An overview of the proposed system. Point cloud of the current assembly ($P^t$) is processed by a graph neural network. The selected action is rewarded by comparing the updated assembly ($P^{t+1}$) with the target $P^T$.}
    %Representation of the assembly in RL setting. Agent extracts features $\mathcal{Z}$ from the state using 3 layered GCN (pink), then, estimates an action using a MLP with 2 hidden layers (blue). Environment stores $\mathcal{X}$ and $\mathcal{A}$ (details in \ref{subsect:prob_def} and \ref{subsect:Representation}). Moreover, it stores Part Poses, Connection Site Information and Part Connection Information in order to update state graph and point cloud representation of assembly.}
    \label{fig:overview}
\end{figure}

\subsection{Overview}

We use DRL (Proximal Policy Optimization \cite{PPO} to be specific) to find a policy $\pi$ for successful assembly of furniture. Following similar studies \cite{4_Learn2Assembly, 3_SingleImage,1_3Dgen}, we use graphs to encode the state of the environment (Sect. \ref{subsect:Representation}) and devise a GNN to obtain a probability distribution over the actions for successful assembly (Fig. \ref{fig:overview} and Sect. \ref{subsect:GNN}). To train the network, we propose a novel reward function (Sect. \ref{sect:reward}) that consists of two measures: Incorrectness and Incompleteness, which are computed by matching $P^t$ and $P^T$. See Fig. \ref{fig:overview} for an overview.

\subsection{Graph Representation of Assembly State}
\label{subsect:Representation}

\begin{figure}[hbt!]
    \centering
    \includegraphics[width=0.85\linewidth]{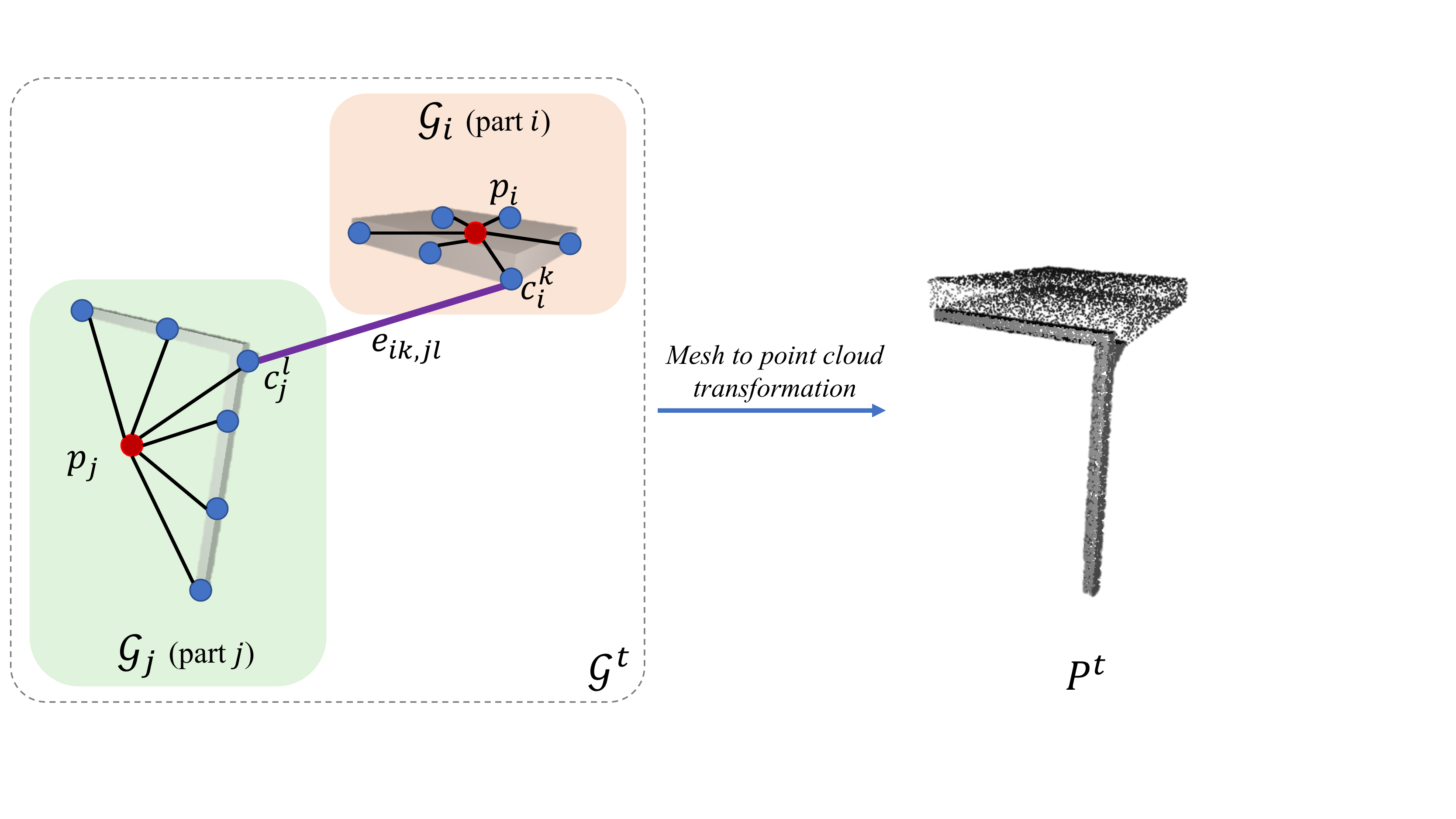}
    \caption{Each part is represented by a graph ($\mathcal{G}_\cdot$), consisting of a part node ($p_\cdot$ -- red circle) and 6 connection nodes ($c_\cdot$ -- blue circle). Parts are attached by drawing edges ($e_\cdot$) between connection nodes. $P^t$ is obtained by sampling the connected parts.}
    \label{fig:representation}
\end{figure}

{\bf The state of the assembly} is represented as a graph. 
%{\bf Part representation.} 
Initially, each of the $N$ parts to be assembled is represented as a separate undirected graph. Specifically,  the $i^{th}$ part is represented as a graph $\mathcal{G}_i=(p_i, c_i^1, c_i^2, .., c_i^6)$ consisting of a {\em part node} ($p_i$) attached to $6$ {\em connection nodes} ($c_i^1, .., c_i^6$), shown as red and blue nodes respectively in Fig.~\ref{fig:representation}. The representation of a part node $p_i$, denoted by $\phi(p_i)$, is the point cloud representation for the part, whereas a connection node $c_i^k$ is represented by a one-hot vector, denoted by $\phi(c_i^k)$. % are associated with the relative pose of the $k^{th}$ connection site with respect to the center of the part. 

{\bf Assembly actions} are represented as a tuple $(\mathcal{G}_i, \mathcal{G}_j, c_i^k, c_j^l)$, where $\mathcal{G}_i, \mathcal{G}_j$ represent the parts, and $c_i^k, c_j^l$ represent the connection sites on these parts. An assembly action would merge $\mathcal{G}_i, \mathcal{G}_j$ by adding an edge, $e_{ik,jl}$, connecting the $k^{th}$ connection site of part $i$ with the $l^{th}$ connection site of part $j$. 

One of the parts, say $\mathcal{G}_i$, is picked as the `seed' and is used to initialize the {\em partial assembly} graph denoted with  $\mathcal{G}^0 = \mathcal{G}_i$. The state of the partial assembly at step $t$, denoted as $\mathcal{G}^{t}$, is updated at every assembly step by merging the graph representations of other parts through edges formed by connections. 

%= \mathcal{G}_p^{t-1} \cup \mathcal{G}_j \cup e_{ik,jl}$, is defined as the connected component containing the `seed' part. 

%.  say part $i$ to be denoted as $\mathcal{G}^0 = \mathcal{G}_i$, where $\mathcal{G}^0$ denotes the state of the partial assembly at step $t=0$. The state of the partial assembly at step $t$, denoted as $\mathcal{G}^{t}$, is defined as the connected component containing the `seed' part. 

At step $t$, an action tuple $(\mathcal{G}_i, \mathcal{G}_j, c_i^k, c_j^l)$  is considered {\em valid} if {\bf (i)}~one of $\mathcal{G}_i$, $\mathcal{G}_j$ is a subset of $\mathcal{G}^t$, the partial assembly, while the other is not a subset of $\mathcal{G}^t$, and {\bf (ii)}~$e_{ik,jl} \not\in \mathcal{G}^t$. All other actions are considered {\em invalid}.

The state of the assembly $\mathcal{G}^t$ is then converted into a representation suitable to be fed to neural networks, as shown in Fig.~\ref{fig:overview}. Specifically, a feature matrix, $\mathcal{X} \in \mathbb{R}^{(N+6N)\times256}$, is used to store a processed representation of the $N$ parts and the $6N$ connections sites. For each part node $p_i$, its point cloud $\phi(p_i)$ is fed into PointNet++ \cite{pointnetplusplus} pretrained on ModelNet \cite{modelnet} to compute a  feature vector $\mathcal{X}_i \in \mathbb{R}^{256}$. The $6N$ connection sites of all $N$ parts are represented as one-hot vectors. For example, connection site ${c_i^k}$ is represented with a $1$ at position $6i + k$. In our study, the size of one-hot vector, $6N$, did not exceed the size of geometric feature vector $256$ and to be compatible with part features, the remaining dimensions are padded with zeros. Finally, the connectivity of the undirected graph $\mathcal{G}^t$ is represented as an adjacency matrix, $\mathcal{A} \in \mathbb{R}^{(N+6N)\times (N+6N)}$.

\subsection{Graph Neural Network}
\label{subsect:GNN}

We constructed a deep network that consists of a graph convolutional subnetwork (GNN), followed by a multi-layer perceptron ($\text{MLP}$):
%\begin{equation}
%\label{eqn:gcn}
%    \mathcal{Z} = f_{gfe}(\mathcal{X}, \mathcal{A})
%\end{equation}
\begin{equation}
\label{eqn:GNN}
    \mathcal{P} = \text{MLP}(\text{GNN}(\mathcal{X}, \mathcal{A})),
\end{equation}
where, as introduced in Sect. \ref{subsect:Representation}, $\mathcal{X}$ is the feature matrix representing the nodes, and $\mathcal{A}$ is the adjacency matrix. For GNN, we used three layers of graph convolution operator (denoted by GCo) from \cite{kipf2017semi} which modulates node features of a part with respect to other parts through the connected nodes of the connection sites: 
\begin{equation}\footnotesize
\label{eqn:gcn}
    \text{GNN}(\mathcal{X}, \mathcal{A}) = \text{ReLU}(\text{GCo}(\text{ReLU}(\text{GCo}(\text{ReLU}(\text{GCo}(\mathcal{X}, \mathcal{A})), \mathcal{A})), \mathcal{A})),
\end{equation}
where $\text{ReLU}$ is a rectified linear unit. $\text{GNN}(\mathcal{X}, \mathcal{A})$ yields $\mathcal{Z}\in \mathbb{R}^{(N+6N)\times256}$ as the processed feature matrix. This feature matrix is `flattened' and provided to $\text{MLP}(\cdot)$ as input.

$\text{MLP}(\cdot)$ is a multi-layer perceptron with two hidden layers to estimate log probabilities of each action:
\begin{equation}
\label{eqn:action}
    \text{MLP}(\mathbf{x}) =   \text{FC}(\text{ReLU}(\text{FC}(\text{ReLU}(\text{FC}(\mathbf{x}))))),
\end{equation}
where $\text{FC}$ is a fully-connected layer.

\subsection{Reward Function}
\label{sect:reward}

Reward is computed using only the point cloud of the fully assembled furniture $P^T$ (as summarized in Alg. \ref{alg:reward}).  Unlike \cite{2_RoboAssembly, 1_3Dgen}, ground truth information about the connections between connection site pairs and relative part poses are not assumed, making the problem setting more practical yet more challenging. 

\begin{algorithm}[hbt!]
\caption{The proposed reward function.}\label{alg:reward}
\scriptsize
\textbf{Input}: $P^t$: Point cloud of partial assembly,\\
    \hspace*{0.6cm} $P^T$: Point cloud of target assembly, \\ 
    \hspace*{0.6cm} $d^{t-1}$: $\COMP$ measure at $(t-1)$.\\
\textbf{Output}: $Reward$: Reward at the end of step, \\ 
    \hspace*{0.8cm} $Termination$: Termination condition.

\begin{algorithmic}
\State $Termination \gets false$
\If {$\mathcal{G}_i,\mathcal{G}_j \nsubseteq \mathcal{G}^t$} \Comment{\textcolor{gray}{\footnotesize Invalid: Neither part is in partial assembly}}
\State $Reward \gets -10$, $Termination \gets true$ 
\ElsIf {$\mathcal{G}_i, \mathcal{G}_j \subset \mathcal{G}^t$} \Comment{\textcolor{gray}{\footnotesize Invalid: Parts are already assembled}}
\State $Reward \gets -1$
\Else %\Comment{\textcolor{gray}{\footnotesize Valid}}
    \State $d^t = \COMP(P^t, P^T)$ \Comment{\textcolor{gray}{\footnotesize Eq. \ref{eqn:comp}}}
    \If{$d^t \geq d^{t-1}$} \Comment{\textcolor{gray}{\footnotesize Completeness $\downarrow$}}
        \State $Reward \gets -10$, $Termination \gets true$ %\Comment{Incomplete action}
    \Else
        \If{$\CORR(P^t, P^T) < n_{th}$} \Comment{\textcolor{gray}{\footnotesize $n_{th}$: threshold}}
            \State $Reward \gets 5$ \Comment{\textcolor{gray}{\footnotesize Correctness $\uparrow$}}
        \Else
        \State $Reward \gets -5$  \Comment{\textcolor{gray}{\footnotesize Correctness $\downarrow$}}
        \EndIf
    \EndIf
\EndIf
\State \Return $Reward, Termination$

\end{algorithmic}
\end{algorithm}

%(i) result in negative reward and unchanged state. For (ii), the state changes but a high negative reward is returned and the episode is terminated. From here on, all discussion on action and reward assumes actions are valid.

{\bf Point cloud of partial assembly}, $P^t$, is generated from the current state of the assembly, represented in  $\mathcal{G}^{t}$ and is updated after every action. To obtain $P^t$, at first, the mesh of the part assembly is obtained; and then a fixed number of points are sampled from this mesh. For each $P^t$, the same number of points are sampled, resulting in a density change after addition of new parts. The affect of the density change is discussed in the following section.

\begin{figure}[hbt!]
    \centering
    \includegraphics[width=0.8\linewidth]{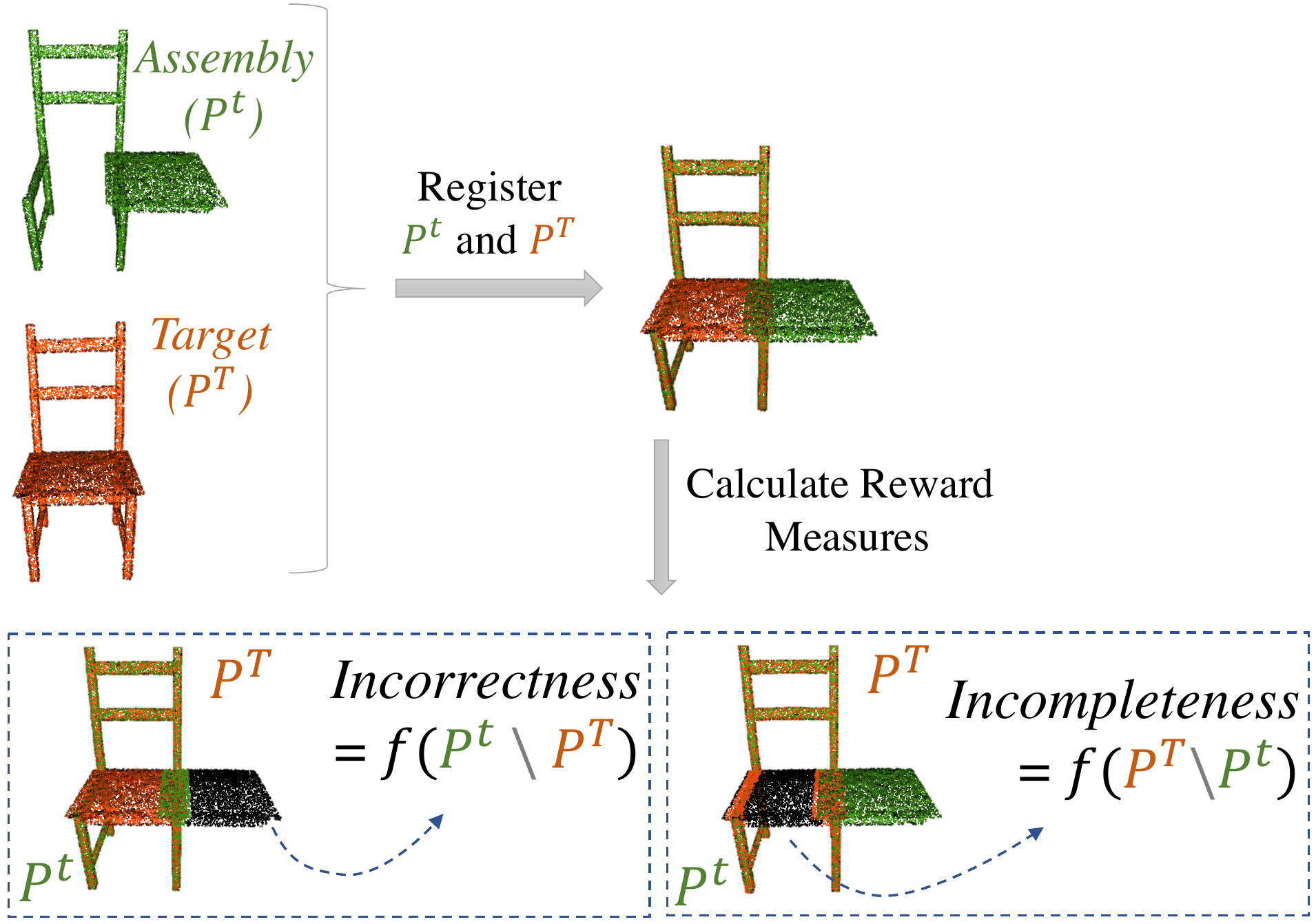}
    \caption{An illustration of the two reward measures proposed in our paper.}
    \label{fig:measurement}
\end{figure}

%The current state of the assembly, represented in  $\mathcal{G}^{t}$, is first used to generate point cloud of the partial assembly, $P^t$, and is updated after every action. 
%Parts connected by agent to the anchor part form the source point cloud $p^S$, which is computed using $\mathcal{C}_{anchor}$. 
The partial assembly, $P^t$, is then compared against $P^T$ using two evaluation measures inspired from Chamfer Distance; namely \textit{incompleteness} and \textit{incorrectness} (Fig. \ref{fig:measurement}). The incompleteness measure is designed to measure the completion progress of the partial assembly towards the final furniture, whereas the incorrectness measure aims to measure the degree of incorrect part assembling. 

In order to discount the effect of the arbitrary pose of the partial assembly during its comparison with the final assembly, Iterative Closest Point (ICP)~\cite{icp} is used for the point cloud registration between $P^t$ and $P^T$. Unassembled parts are excluded during the registration step and metric computation that follows it. %Moreover, the transformation if the assembly action 

\begin{figure}[hbt!]
    \centering
    \includegraphics[width=0.99\columnwidth]{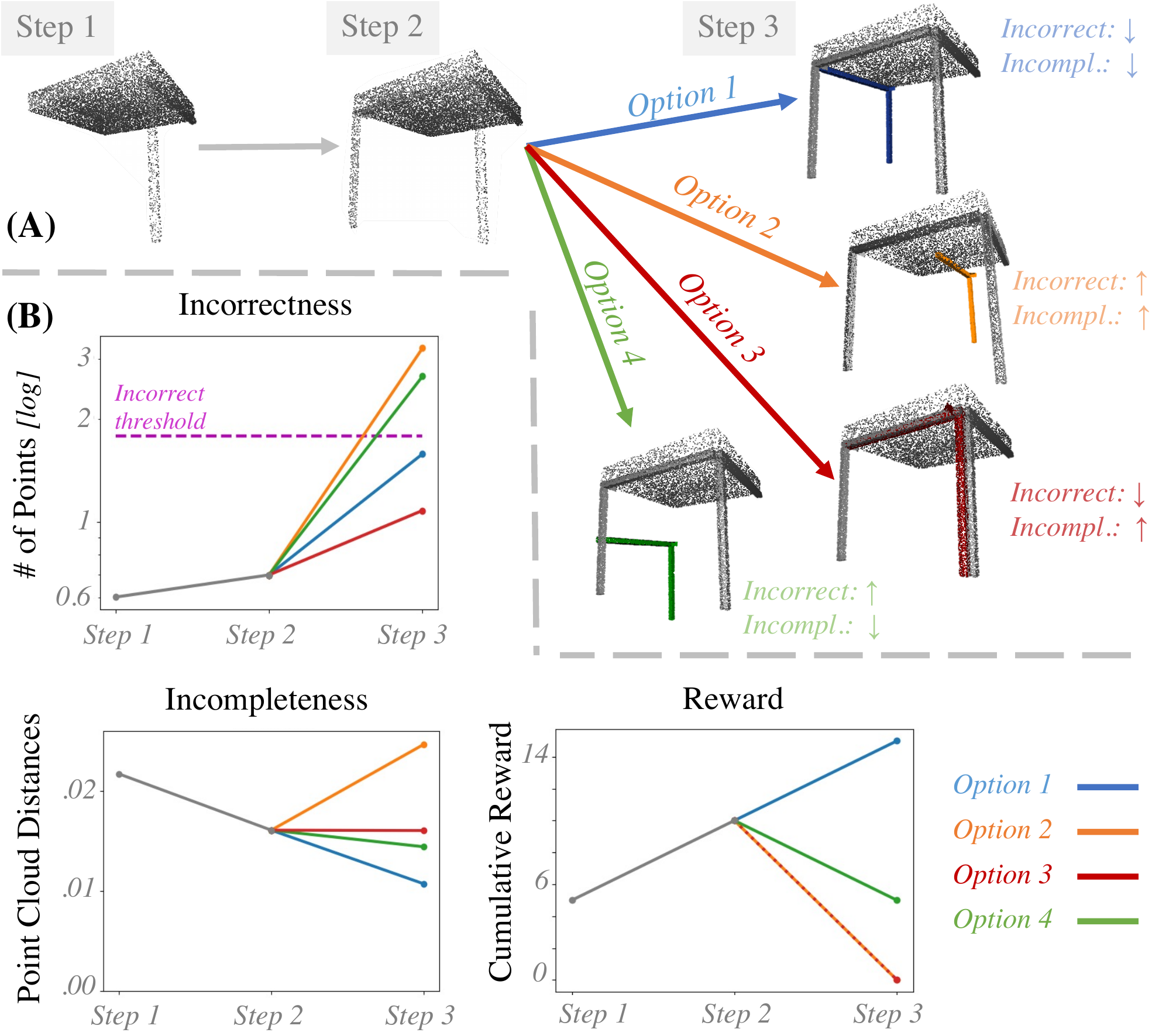}
    \caption{An illustration showing how the proposed measures and the reward values change over time. (a) Assembly of a table with three steps. (b) How different assembly actions (options in (a)) affect the measures.}
    \label{fig:reward}
\end{figure}

%Unlike bidirectional CD, these metrics are computed unidirectional. However, computing metrics directly is not appropriate. Note that $p^S$ and $p^T$ have different transformations since assembly started with random poses. Consequently, Iterative Closest Point (ICP) \cite{icp} is incorporated in order to register $p^S$ and $p^T$. As mentioned above, an anchor part is selected before training, which has two ICP originated motivations: (1) point cloud of assembled parts instead of point cloud of scattered parts and (2) proper selection of anchor (e.g. part that has largest surface area) lead better point cloud registration.

{\bf The incompleteness measure} is defined as the average distance from $P^T$ to $P^t$ as:
\begin{equation}
\label{eqn:comp}
    \COMP(P^T, P^t) = \dfrac{1}{|P^T|} \sum_{x\in P^T} \min_{y\in P^t}||x-y||^2_2,
\end{equation}
where $x \in P^T$ and $y \in P^t$, are points in the point clouds of the fully and partially assembled furniture. The calculation is normalized with the number of points in the fully assembled furniture, denoted with $|P^T|$. The measure is expected to decrease as the assembling progresses correctly (see Fig. \ref{fig:reward}), but is not expected to drop to complete zero.

%To compute the distance for a point $x$ in $p^T$, the closest pair of $x$ in $p^S$ is used. Initially $p^S = p^{anchor}$, thus, all points in $p^T$ has the closest pair on the $ p^{anchor}$. 
A correct assembly action would certainly reduce the average distance between paired points of the two point clouds. An incorrect assembly action, however, may result in two possible changes on the measure. If the incorrectly assembled part hampers ICP-based registration in the previous step, it would lead to an increase in the measure, as expected. However, if the assembled part fails to hamper ICP-based registration, then the points in $P^T$ will still be paired to the same or close-by points in in $P^t$, not causing a noticable increase in the measure. Moreover, the assembly of a new part into the partial assembly would increase the surface area and would make the point cloud sampled from the mesh sparser in the areas that match $P^T$. As a result, the measure may even decrease unexpectedly.   

%in which the negative reward is returned and the episode is terminated. Nevertheless, the wrong action may decrease $\COMP$ when point cloud registration is not affected. This reveals the need for correctness.

%An action towards the target assembly decreases $\COMP$ by bringing together parts in such a way that their assembled point clouds are close to $p^T$. On the other hand, a wrong action has three possible effects. It may increase $\COMP$ by worsening point cloud registration in which case a negative reward is returned and the episode is terminated. Moreover, a wrong action may decrease $\COMP$ when point cloud registration is not affected, due to closest points in the partial assembly point cloud, $P^t$, not changing. Lastly, since each part is sampled with a fixed number of points, attaching a new part makes $P^t$ less dense and with an incorrect step, the closest points may change, slightly increasing the $\COMP$. Therefore, an additional measure for correctness of a step is needed. 

The unexpected behavior of the incompleteness measure is compensated by 
{\bf the incorrectness measure}, which is defined as:
\begin{equation}
\label{eqn:corr}
    \CORR(P^t, P^T) = \sum_{x\in P^t} \mathbb{1} \left(\min_{y\in P^T}||x-y||^2_2 > d_{th}\right), 
\end{equation}
which depends on the cardinality of points in $P^t$ whose distances to the closest pairs in $P^T$ are higher than the threshold $d_{th}$. Assuming correct point cloud registration, if an action leads to a correct assembly, $\CORR$ is low. If an action yields an incorrect assembly, some 3D points must be distant from the target by a threshold $d_{th}$. If there exists such $\CORR$ points more than a constant number $n_{th}$, a negative reward is returned. 
%% Erol: Asagidaki cumleyi Experiment kismina aktardim.
%In our study, we picked $d_{th} = 1.5 cm$, $n_{th} < 200$ and $|p^S|, |p^T| = 10000$ 

\iffalse
\begin{itemize}
    \item Partial (parts connected to a main object) point cloud
    \item ICP
    \item Distance Computation and How these distances are used
    \item Two thresholds and why they are used
\end{itemize}
\fi

There are two types of invalid actions. For actions that do not change the state, a negative reward is given and for the actions that do not connect a part to the partial assembly $\mathcal{G}^{t}$, a high negative reward is returned and the episode is terminated. From here on, all discussion on action and reward assumes actions are valid.

\subsection{Learning a Policy for Assembly}

Having defined a representation for state (Sect. \ref{subsect:Representation}), a neural network (Sect. \ref{subsect:GNN}), and a reward (Sect. \ref{sect:reward}), we  use an off-the-shelf DRL method, namely Proximal Policy Optimization (PPO) \cite{PPO}, in an actor-critic style to learn a policy $\pi$ for the assembly actions and a critic function for computing advantages of the actions. The PPO algorithm is chosen over other other actor-critic methods due to its better performance \cite{PPO}.
\section{Experimental Setup}
\label{sect:experiments}

% \textit{Although there is no previous work studying the same problem setting as ours, we formulate three strong baselines inspired by previously published works on similar task domains and demonstrate that our method outperforms baseline methods by significant margins}
%In this section, we provide the implementation and training details, the quantitative results of the proposed method, and a comparison with two baseline models \& similar distance-based reward implementations.

%\subsection{Implementation, Training and Evaluation Details}

A {\em furniture library} of 11 different furniture models, consisting of STL mesh files specifying the shape of the parts, and an XML file in MuJoCo Model specifying the relative poses of the connection sites and the connections, are imported from the IKEA Furniture Assembly Environment \cite{lee2021ikea}. For each part, new connection sites were manually added to increase the number of sites to 6 per part. This modification was made (i) to have a fixed-size representation for each part to be fed into the neural network, as well as (ii) to make the assembly task much more challenging.  For each furniture model the `seed' is selected as the largest part.

The \textit{Combinatorial Complexity} of the assembly learning problem for 11 furniture is defined as the ratio of the number of correct action sequences to the all possible sequences. In our study, for an $N$-part furniture with 6 connection sites per part, the number of possible action sequences is:
\begin{equation}
    \label{eqn:comb}
    \prod_{i=1}^N \binom{i}{1} \cdot \binom{N-i}{1}\cdot\binom{6}{1}\cdot\binom{6}{1}.
\end{equation}
%that is the number of sequence of actions that can be selected for each step to finish an assembly. 
The number of correct action sequences that would yield a successful assembly is 
computed taking into consideration non-unique parts and connections (such as the four indifferent legs of a table, that can be installed on either end). 

Table~\ref{tab:exp1} shows the combinatorial complexity of the furniture in the library. 
For instance, for the Lack [\textit{table}] (see Fig. \ref{fig:visual_results}), which consist of one table top and four identical legs which can be installed on either end, the  number of correct action sequences computed as:  
%\begin{equation*}
%    (\binom{4}{1}\cdot\binom{4}{1}\cdot\binom{2}{1})\cdot(\binom{3}{1}\cdot\binom{3}{1}\cdot\binom{2}{1})\cdot(\binom{2}{1}\cdot\binom{2}{1}\cdot\binom{2}{1})\cdot\binom{2}{1}
%\end{equation*}
\begin{equation*}
    \prod_{i=1}^4 \binom{i}{1}\cdot\binom{i}{1}\cdot\binom{2}{1},
\end{equation*}
where, at each step, one of the correct and empty connection sites on the table top and one of the legs that is not already connected are selected as a pair to connect. Additionally, since legs have symmetry, one of connection sites are selected to connect.  

%We see from the table that the problem is highly challenging, yet the proposed method with our novel reward function is able to find the correct assembly sequences in a large space and it can assemble different types of furniture.

The relative poses of the connection sites and the connections between the parts were not used by the \METHOD. Instead this information are used both as the ground-truth information against which \METHOD~is evaluated, as well as to build alternatives to compare its performance.

%are taken from the IKEA Furniture Assembly Environment. The objects include STL mesh files for shape information of the parts and a XML file in MuJoCo Model (MJCF) . 

%It includes the part information, the target assembly and information about the connection sites. Small adjustments are done to add more connection sites to each part for all objects. The target point cloud by sampling the mesh obtained using the XML files. The target point cloud does not include any part related information. Each part is stored as a mesh. When the reward will be computed, meshes of the the main part and the parts connected to the main part are transformed using the pose information and then sampled to obtain the source point cloud.

%Details of the gym environment, how the actions work and how changes in the states are represented.

%\textbf{Environment}: 
\textbf{Training \& Testing Details}: The training and testing are implemented and conducted within the Gym  Reinforcement Learning environment \cite{gym}. The graph, mesh and point cloud representations of the parts, as well as the partially and fully assembled furniture are stored, updated and generated with custom-built extensions.  Specifically, the extension supported (i)~the generation of point cloud of a part or partially assembled furniture from its mesh and connectivity information, (ii)~the update of the graph representations as a result of an action, and (iii)~reward computation, as described above.   During training, initial pose of the parts are set randomly once and remained the same for all the episodes. 
%That is the initial poses are fixed throughout all training. 

%Moreover, to manipulate the connected parts as one rigid body, the connection sets of the parts are also stored by the environment. Lastly, for the actions that fit both correctness ($\CORR$) and completeness ($\COMP$) measures, the transformation computed by the ICP algorithm is used to manipulate the parts of the partial assembly.

%\textbf{Training Details}: 
%Not forget to mention training same model for multiple times (mean and var).
The training and testing experiments are carried out within the Rl-baselines3-zoo \cite{rl-zoo3} framework. The PPO \cite{PPO} algorithm was used with the default parameters, and the maximum number of steps were set as 10,000. 
During the experiments, the distance threshold ($d_{th}$) is set as $1.5cm$ and the number of points threshold ($n_{th}$) is chosen between 10 and 200 for different furniture.

%For training with the proposed method, default parameters of the PPO algorithm are used with the maximum number of steps  chosen as 10,000. 
%In our study, we picked $d_{th} = 1.5 cm$, $n_{th} < 200$ and $|p^S|, |p^T| = 10000$ 

%We trained all the methods with 11 different furniture models, and evaluated the models with the number of correct assemblies. For the models that fail to obtain the correct assembly, we evaluate the ratio of the correct connections.

\textbf{Evaluation Measures}: The performance of assembly learning methods is evaluated with two measures, which use the ground-truth information included in the furniture models: (1) At the part connection level, we define SR$_\text{con}$, \textit{connection success rate}, as the  ratio of correct connections done by the agent to the  number of total correct connections. In Table \ref{tab:exp1}, connection success rate for each furniture, and in Table \ref{tab:exp_2}, connection success rate for all furniture are shown. (2) At the furniture assembly level, we use SR$_\text{a}$, {\em furniture assembly success rate}, as the ratio of correctly assembled furniture to the total number of furniture.

%The \textit{Connection Success Rate}, denoted as  SR$_\text{c}$, is defined as the ratio of correct connections to the  number of connections on a furniture. The \textit{Assembly Success Rate}, denoted as SR$_\text{a}$), is defined as the ratio of correct assembled furniture to all furniture.

%In order to illustrate the complexity of the problem, we also provide in the table an analysis of 

\section{Experiments and Results}

\subsection{Baseline Models}\label{sect:baselines}
We consider two new reward functions as baselines, which provide more supervision using the correct connection pairs required for each furniture are proposed: {\bf BL1}: Provide positive reward (+5) for each action that connects the correct pairs, and negative reward (-5) otherwise. {\bf BL2}: 
    Provide positive reward (+5) if the furniture is correctly assembled at the end of the episode, and negative reward (-5) otherwise. 

\subsection{Experiment 1: Furniture Assembly with \METHOD}
\label{exp:1}
%Train model for different models for n nodes where n is the number of parts. This requires different network architecture (input size, graph size)
The last column of Table~\ref{tab:exp1} shows the success rate (SR$_\text{con}$) for \METHOD. The results show that \METHOD\ can learn assembly sequences effectively.

%Since furniture have different number of parts, the state representation and action space size change with each furniture. To evaluate our method, we measure the correct number of connections as the success rate. %We conducted Experiment 1 with the reward function consisting of correctness (Eq. \ref{eqn:corr}) and completeness (Eq. \ref{eqn:comp}) measures combined in our reward function (Alg. \ref{alg:reward}) provided at each step.

\begin{table}[H]
    \centering
    \caption{Combinatorial complexity (Eq. \ref{eqn:comb}) of furniture in the library and individual connection rate for \METHOD\  (Alg.~\ref{alg:reward}). \label{tab:exp1}}
    \scriptsize
    \begin{tabular}{|c|c|c|c|}
         \hline
         \textbf{Furniture} & \textbf{\# parts} & \makecell{\textbf{Combinatorial} \\ \textbf{Complexity}} & $\mathbf{SR_{con}}$\\
         \hline \hline
         Agne \hfill \textit{[Chair]} & 3 & $2 \ / \ 518$ & 2/2 \\
         \hline
         Bernhard \hfill \textit{[Chair]} & 3  &$2 \ / \ 5184$ & 2/2\\
         \hline
         Swivel \hfill \textit{[Chair]} & 3  & $1 \ / \ 5184$ & 2/2\\
         \hline 
         Bertil \hfill \textit{[Chair]} & 5  & $12 \ / \ (0.9\times10^9)$ & 4/4\\
         \hline
         Ivar \hfill \textit{[Chair]} & 5  & $136 \ / \ (0.9\times10^9)$ & 4/4 \\
         \hline
         \hline
         Mikael \hfill \textit{[Table]} & 4 & $8 \ / \ (1.7 \times 10^6)$ & 2/3 \\
         \hline
         Klubbo \hfill \textit{[Table]} & 5 & $8 \ / \ (0.9\times10^9)$  & 4/4\\
         \hline
         Lack \hfill \textit{[Table]} & 5  & $9216 \ / \ (0.9\times10^9)$ & 4/4\\
         \hline
         Tvunit \hfill \textit{[Table]} & 5  & $104 \ / \ (0.9\times10^9)$ & 4/4\\
         \hline \hline
         Ivar \hfill \textit{[Shelf]} & 6  & $14400 \ / \ (0.8\times10^{12})$ & 5/5\\
         \hline
         Liden \hfill \textit{[Shelf]} & 11  & $2\times10^7 \ / \ (4.8\times10^{28})$ & 1/10\\
         \hline
         
    \end{tabular}
    
\end{table}

%\BB{Elimizde random initli best modelin de trainingleri var}

Although the results in Table \ref{tab:exp1} are  promising for a method with weak supervision, the model fails to assemble certain products; furniture with small parts ({\em Liden [Shelf]}), furniture not containing distinct parts ({\em Liden [Shelf]}) and furniture with isotropic parts ({\em Mikael [Table]}). Small parts are problematic because their mis-attachment may be missed by our measures owing to their relatively small size. % Adjusting the thresholds helps to detect the correctness of even the smaller parts.
Furniture without a distinct part ({\em Liden [Shelf]}) can prevent ICP from correctly registering the point clouds, consequently affecting the reward measures. Isotropic parts (e.g. in  {\em Mikael [Table]}) can be registered very well by ICP, though connection sites may be flipped, which is not captured by our measures, which may lead to incorrect attachment of consecutive parts. %is the orientation of the connected parts and the placements of the connection sites. The point cloud of the connected objects can be registered correctly and visually seems correct, however connection sites end up in wrong relative position, resulting in the other parts to be connected wrong.

%\BB{Aşağıdaki paragraf anlaşılır olmadı onun için bir figür koyuyorum. Aşağıdaki şekilde bacakları birleştiren uzun destek paçası buna bir örnek. Bu parçayı bağlamak için conn site masanın iç tarafına doğru olmalı. Ancak agent bazen conn site dışa bakarken ancak ara parçanın rotasyonu içe olacak şekilde birleştirmeyi öğreniyor. (Yani ara parça olması gereken yerin çoğunu işgal ediyor ancak dışarıdan başladığı için bağlanacağı diğer bacağa tam denk gelmiyor arada minik bir boşluk oluyor.) Burada bacağın içinden geçiyor ve collision check olsa valid bir hareket olmayacak. Bu paragrafı kaldırabiliriz ya da daha güzel açıklanabilirse tutabiliriz.}
%\begin{figure}[H]
%    \centering
%    \includegraphics[width=0.5\linewidth]{Figures/Capture.PNG}
%    \caption{Caption}
%    \label{fig:my_label}
%\end{figure}

%For the {\em Shelf Ivar}, the model learns how to connect 3 parts but did not learn to connect the last 2 parts. After connecting 3 parts, the model selects the same actions again, getting stuck at the sub-assembly. In this case increasing the training steps or learning rate can improve the learning.

%\OA{Çok yer kaplayabilir ama modellerin bir kısmının (özellikle hata yapılanların) görsellerini koyabiliriz (pointcloud ya da mesh olarak)}

%\subsection{Experiment 2: Robustness of Assembly}
%\OA{Bu bölümdeki deneyleri yapamadık} 
%Train model for different models for fixed number of parts. This fixes model architecture but increases complexity. For eg. If selected n = 10
% and furniture has 6 parts. 4 remaining part will be random parts from another furniture.
 
\subsection{Experiment 2: Ablation Study}

Now we evaluate the contributions of different design choices for our reward function. Table \ref{tab:exp_2} suggests that using the incorrectness and incompleteness measures provides the best performance. We observe that Chamfer Distance (or using its difference between consecutive steps -- delta Chamfer Distance\footnote{At the start of an episode and after each step, Chamfer Distance is computed, and the difference between the distances of consecutive steps are used as the reward. The hypothesis is that this delta Chamfer Distance better reflects the progress (the effect of an action) by capturing the decrease in the distance.}) is not able to provide sufficient training signals for assembling many of the furniture. 

\begin{table}[H]
    \centering
    \caption{Ablation study on the reward function. Den.: Reward at each step, Spa.: Reward at episode end. CD: Chamfer Distance, $\delta$-CD: Difference in Chamfer Distances.}
    \label{tab:exp_2}
    \scriptsize
    \begin{tabular}{|c|c|c|c|c|c|c|c|}
        \hline
        \multicolumn{2}{|c|}{\textbf{Reward Freq}} & \multicolumn{4}{|c|}{\textbf{Reward Type}} & \multicolumn{2}{|c|}{\textbf{Measures}}\\
        \hline
        \textbf{Den.} & \textbf{Spa.} & $\mathbf{\CORR}$ & $\mathbf{\COMP}$ & \textbf{CD} & $\delta-$\textbf{CD}& $\mathbf{SR_a}$ & $\mathbf{SR_{con}}$\\
        \hline\hline 
       \ding{51} & & & & \ding{51}& &  2/11 & 15/44\\ \hline
       \ding{51}& & & & & \ding{51} &   3/11 & 15/44 \\ \hline
       \ding{51} & & \ding{51} & & & &  {4/11} & {20/44}\\ \hline
       \ding{51} & &   & \ding{51}& & &  {1/11} & {9/44}\\ \hline
       \ding{51} & & \ding{51} & \ding{51}& & &  \textbf{9/11} & \textbf{34/44}\\ \hline
        & \ding{51} & \ding{51} & \ding{51}& & &  0/11 & 0/44 \\ \hline
%        &\ding{51} & & &  & \ding{51}&  2/11 & \\
%       \hline
%        & \ding{51}& & &  & \ding{51}&  \textbf{10/11} & \textbf{35/44} \\
%        \hline
    \end{tabular}  
\end{table}

\subsection{Experiment 3: Comparison with Baselines}

We now compare the best setting of our method with two strong baselines (BL1 and BL2) that use strong supervision (Sect. \ref{sect:baselines}). Table \ref{tab:comparison_with_baseline} shows that the proposed weak supervision at each assembly step provides comparable performance to using strong supervision at each step, and it performs better than using strong supervision at the end of an episode.

\begin{table}[H]
    \centering
    \caption{Performance comparison of \METHOD\ with BL1 and BL2. 
    \label{tab:comparison_with_baseline}}
    \scriptsize
    \begin{tabular}{|c|c|c|c|c|c|c|}
    \hline
        & \multicolumn{2}{c|}{\textbf{Reward Freq}} &\multicolumn{2}{c|}{\textbf{Supervision}} & \multicolumn{2}{c|}{\textbf{Measures}}\\
        \hline
        \textbf{Method}  & \textbf{Den.} & \textbf{Spar.} & \textbf{Strong} & \textbf{Weak} & $\mathbf{SR_a}$& $\mathbf{SR_{con}}$\\
        \hline \hline
        Strong Sup. (BL1) & \ding{51}    &           & \ding{51}  & & \textbf{10/11}& \textbf{35/44} \\ \hline
        Strong Sup. (BL2) &              & \ding{51} & \ding{51}  & & 2/11 & 8/44 \\ \hline
        %\hline\hline
        %Weak Sup. [Ours] & \ding{51}   & & \ding{51}   &           & 4/11 & 20/44\\
        %Weak Sup. [Ours] & \ding{51}   & &             & \ding{51} & 1/11 & 9/44\\
        {[\METHOD} & \ding{51}   & &   & \ding{51} & \textbf{9/11} & \textbf{34/44}\\
        \hline
    \end{tabular}
\end{table}

\subsection{Experiment 4: Qualitative Results}

\begin{figure}
    \centering
    \includegraphics[width=0.7\linewidth]{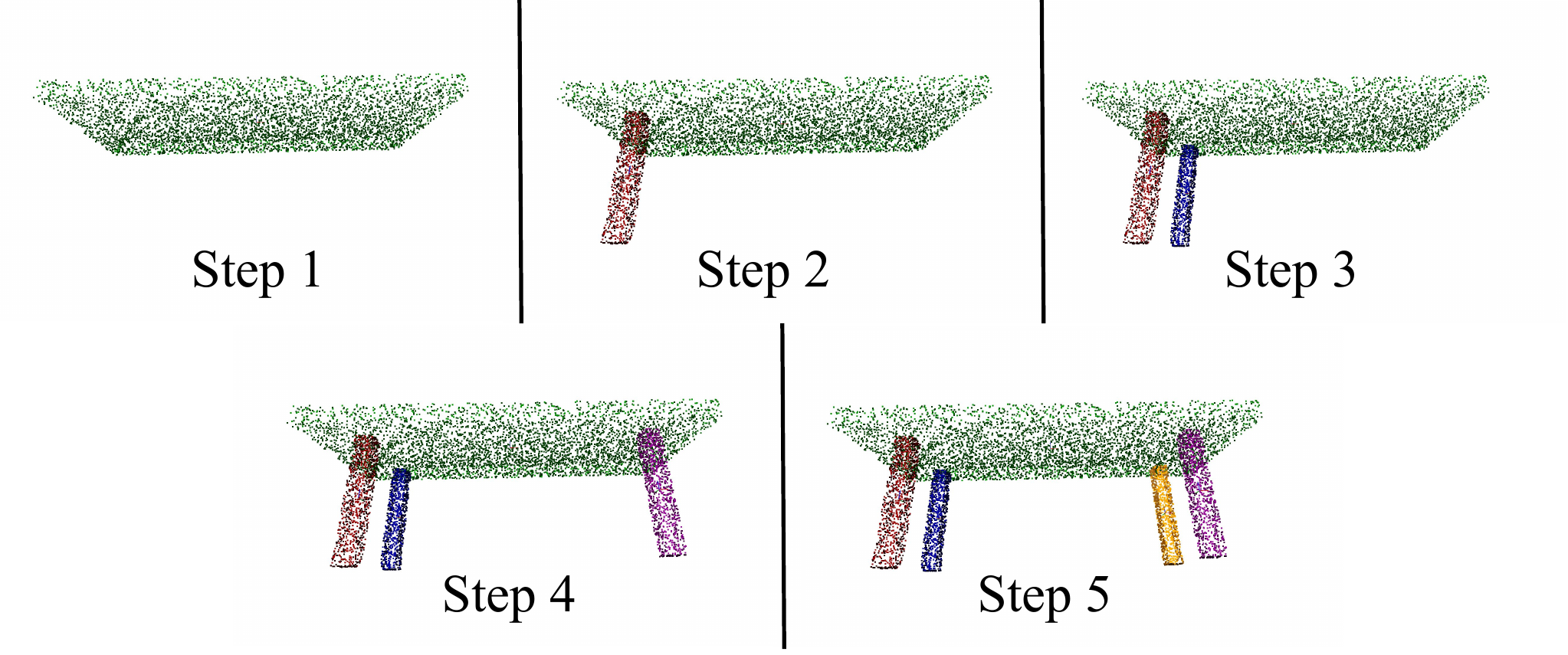}
    \caption{Visualization of the Lack [\textit{Table}] assembly steps. See the supplementary material for more examples.}
    \label{fig:visual_results}
\end{figure}

In Fig. \ref{fig:visual_results}, we provide snapshots from an assembly. In the video provided as supplementary material, we provide more examples, including also failure cases.
\section{Conclusion}

In this paper, we studied whether furniture assembly can be learned only using weak supervision, namely the 3D model of the assembled product. To this end, we proposed a novel reward function consisting of an incorrectness and an incompleteness measure that are calculated by matching the point cloud of the current assembly with the target model. We showed that our novel reward function is able to train a graph convolutional network to assemble various furniture. 

Despite the promising results, our work can be extended in many ways. First of all, our reward function may fail to capture attachment of small parts or isotropic parts, and it can be improved by designing better incorrectness and incompleteness measures. Moreover, instead of using connection sites to assemble parts, the action space of the policy can be changed for the agent to estimate geometric relations between the parts. Lastly, a robotic controller can be used for part manipulation to incorporate robot arm constraints while learning to assemble furniture.

%\subsection{Future Work}
%\begin{itemize}
%    \item Using simulation information to enhance assembly state representation 
%    \item Using learned assembly policy in a collaborative setup
%    \item Causality - Detecting Wrong actions in the process and correcting them
%    \item Learning a robotic controller for assembly using learned assembly policy
%\end{itemize}

%\section*{Acknowledgment}
%This work is partially supported by TÜBİTAK through projects ``ÇIRAK: Compliant Robot Manipulator Support for Assembly Workers in Factories" (117E002) and ``KALFA: New Methods for Assembly Scenarios with Collaborative Robots" (120E269).

\bibliographystyle{IEEEtran}
\bibliography{references}

\end{document}